\documentclass[10pt,twocolumn,letterpaper]{article}

\usepackage{cvpr}
\usepackage{times}
\usepackage{epsfig}
\usepackage{graphicx}
\usepackage{amsmath}
\usepackage{amssymb}

\usepackage[pagebackref=true,breaklinks=true,letterpaper=true,colorlinks,bookmarks=false]{hyperref}

\usepackage{times}
\usepackage{epsfig}
\usepackage[acronym]{glossaries}
\usepackage{acronym}
\usepackage{enumitem}
\usepackage{comment}

\usepackage{multirow}
\usepackage{caption}
\usepackage{subcaption}
\usepackage{sidecap}
\usepackage{wrapfig}
\usepackage{tabularx, booktabs, ragged2e}
\usepackage{adjustbox}

\usepackage{makecell}
\usepackage{gensymb}
\usepackage{xr}
\usepackage{lipsum}  
\usepackage{arydshln}
\usepackage{mathrsfs}
\usepackage{caption}
\acrodef{m2fpa}[M$^2$FPA]{Multi-yaw Multi-pitch high-quality database for Facial Pose Analysis}
\acrodef{3dmm}[3DMM]{3D Morphable Model}

\acrodef{gan}[GAN]{generative adversarial network}
\acrodef{cnn}[CNN]{Convolutional neural network}
\acrodef{sffgan}[SF-GAN]{SuperFront GAN}
\acrodef{ssim}[SSIM]{Structural Similarity Index}
\acrodef{d}[D]{discriminator}
\acrodef{g}[G]{generator}
\acrodef{lr}[LR]{low-resolution}
\acrodef{hr}[HR]{high-resolution}
\acrodef{sr}[SR]{super-resolution}
\acrodef{si}[SI]{single-image}
\acrodef{mi}[MI]{multi-image}
\acrodef{fr}[FR]{facial recognition}

\cvprfinalcopy 

\def\cvprPaperID{10302} 

\ifcvprfinal\fi
\begin{document}

\title{SuperFront: From Low-resolution to High-resolution Frontal Face Synthesis}

\author{Yu Yin, Joseph P. Robinson, Songyao Jiang, Yue Bai, Can Qin, Yun Fu\\
Department of Electrical and Computer Engineering\\
Northeastern University, Boston, MA\\
{\tt\small \{yin.yu1, robinson.jo, jiang.so, bai.yue, qin.ca\}@northeastern.edu, yunfu@ece.neu.edu}
}

\maketitle

\begin{abstract}
    Advances in face rotation, along with other face-based generative tasks, are more frequent as we advance further in topics of deep learning. 
    Even as impressive milestones are achieved in synthesizing faces, the importance of preserving identity is needed in practice and should not be overlooked. Also, the difficulty should not be more for data with obscured faces, heavier poses, and lower quality. Existing methods tend to focus on samples with variation in pose, but with the assumption data is high in quality. We propose a generative adversarial network (GAN) -based model to generate high-quality, identity preserving frontal faces from one or multiple low-resolution (LR) faces with extreme poses.
   Specifically, we propose \ac{sffgan} to synthesize a high-resolution (HR), frontal face from one-to-many LR faces with various poses and with the identity-preserved. We integrate a super-resolution (SR) side-view module into \ac{sffgan} to preserve identity information and fine details of the side-views in HR space, which helps model reconstruct high-frequency information of faces (\ie periocular, nose, and mouth regions). 
   Moreover, \ac{sffgan} accepts multiple LR faces as input, and improves each added sample. We squeeze additional gain in performance with an orthogonal constraint in the generator to penalize redundant latent representations and, hence, diversify the learned features space. Quantitative and qualitative results demonstrate the superiority of \ac{sffgan} over others.
\end{abstract}

\acresetall
\section{Introduction}
Face-based generative tasks (\eg face rotation~\cite{hu2018pose,huang2017beyond,tran2017disentangled,yin2020dualattention}, hallucination~\cite{bulat2018superfan,chen2018fsrnet,yuyin:landmarks:2020}, and attribute editing~\cite{choi2018stargan,he2019attgan}) have gained more of the spotlight in research communities with the advancement of deep learning. Even still, the practical significance of identity-preservation is frequently overlooked, which is especially a challenge for faces with larger pose and lower quality. Recently, some progress has been made to synthesize frontal faces with large pose variations~\cite{huang2017beyond,li2019m2fpa,qian2019unsupervised}.
However, existing methods focus on faces with large poses, while assuming images are high in quality - previous attempts lose identity information when learning a highly non-linear transformation that maps spaces of \ac{lr} side-views to \ac{hr} frontal-views. 

Both low quality inputs and large pose discrepancy between views make the frontalization problem challenging. Modern-day models usually aim to solve either of the challenges (\ie face super-resolution (SR)~\cite{chen2018fsrnet,bulat2018superfan} or large pose face frontalization~\cite{huang2017beyond,yin2020dualattention}), but then breakdown when faces are both low quality and large poses. To overcome these barriers, simultaneously, we proposed \ac{sffgan}. Hence, \ac{sffgan} synthesizes \ac{hr} and identity-preserved frontal face from an \ac{lr} face with an arbitrary pose. For this, a SR module is integrated into SF-GAN to preserve the identity for low-quality of images. Following this, 
a patch loss is introduced
to solve large pose discrepancies by learning a precise non-linear transformations from LR side-views to HR frontal. Furthermore, we believe that the information in a single LR image under extreme poses is limited (\ie insufficient for synthesizing accurate frontal faces in extreme cases). Hence, we designed the model to accept one-to-many inputs - each added sample further improve the results. A discussion of the three challenges, along with the proposed solutions, is presented for each.


\begin{figure}[!t]
    \centering
    \includegraphics[trim=0.5in 0.08in 0.5in 0in,clip,width=\linewidth]{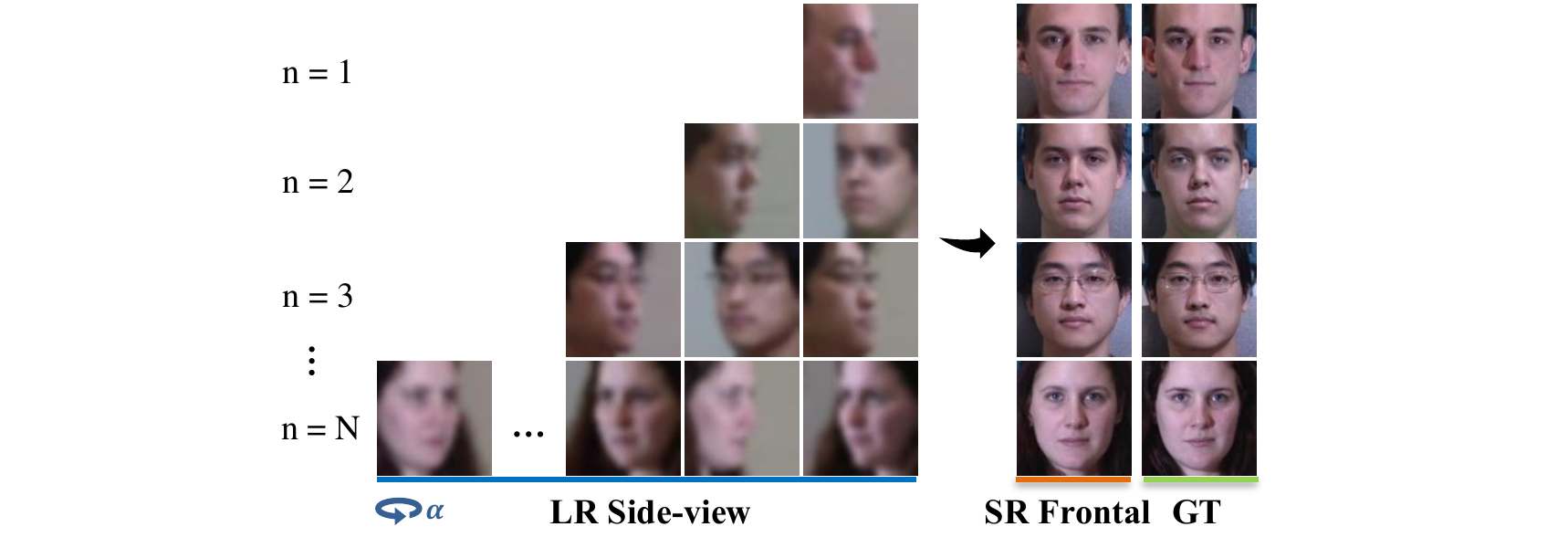}
    \caption{\textbf{Single-to-many input capability.} SF-GAN takes $N$ LR, side-view faces, of arbitrary angle ($\alpha$), as input. The proposed synthesizing high-quality frontal faces from one input, then improves with more added.}
    \label{fig:page1}
\end{figure}

Existing face frontalization methods~\cite{hu2018pose,huang2017beyond,li2019m2fpa,yin2017towards} tend to set the generator as an encoder-decoders with skip connections (\ie U-Net~\cite{ronneberger2015u}). This preserves low-frequency information (\ie shape and uniqueness of objects) by skip connections, while cascades of convolutional-layers learn high-frequency features. However, precise low-frequency information is lost when faced with \ac{lr} inputs. Thus, U-Net architectures lead to inaccuracies at inference in such cases (\eg blurred structures and loss of identity). To properly preserve details and the subject identity of \ac{lr} face, we leverage a \ac{sr} module in parallel to better provide precise low-frequency information and high-frequency details.
The effectiveness of the proposed joint-learning scheme is met with improved quality of high-frequency content, while preserving the identity of the subject. 
To the best of our knowledge, we are the first to address the problem of rotating profile faces and \ac{sr} jointly, and such that the tasks compliment one another.


Another challenge in frontal face synthesis is the highly non-linear transformation from side-to-front view due to large pose discrepancy, leading to imprecise facial structures at inference. 
Previous works~\cite{huang2017beyond,li2019m2fpa,tran2017disentangled} usually use pixel-level (\eg L1 or L2), identity, and adversarial losses to learn mappings between views. However, models trained such losses typically have low confidence in differentiating structural information~\cite{huang2017beyond,qin2019basnet}. 
To capture detailed facial structures as well as identity information of the subject, we incorporate an patch-level loss into the commonly used loss set (\ie pixel loss, identity loss, and adversarial loss), and hence, reinforce the model to pay more attention to image structures (\ie the edge and shape of facial components), (\ie \ac{ssim}~\cite{wang2004image}). 
Different from existing works, we
adds structure-level knowledge in the form of complimentary information provided on the patch-level, showing a significant boost in the final result. 
We show the effectiveness of the patch-based loss in ablation study.

Moreover, synthesizing \ac{hr} and identity preserving frontal views from a single image is often difficult due to extreme poses in \ac{lr} faces. In many real-life scenarios (\eg surveillance system), there are multiple images per subject that can be used in a complimentary fashion to further improve the synthesis~\cite{tran2017disentangled}. However, most existing face frontalization algorithms only handle one image at each time. 
To further boost the quality of the face, we extend our model to accept multiple faces as input (Fig.~\ref{fig:page1}). Since all generators in the proposed model share the same weights, the input image could have arbitrary poses.
Instead of employing naive fusion methods (\eg image- or feature-level concatenation~\cite{reddy2016concatenation}, or feature-level summation~\cite{tran2017disentangled}), we propose using orthogonal regularization in our \ac{gan}-based model for optimal training and to learn features of broader span~\cite{2018arXiv181009102B}. To the best of our knowledge, we are the first to introduce this in training a \ac{gan}-based model. Namely, \ac{sffgan}.

In summary, we make the following contributions:
\begin{enumerate}
    \item To our best knowledge, we are the first to tackle the challenge of tiny face frontalization by proposing a multi-tasking model which learns the frontalization and face super-resolution collaboratively.
    \item We introduce a patch-based loss to capture facial structures and learn a precise non-linear transformation between \ac{lr} side-view and \ac{hr} frontal-view faces.
    \item We extend one-to-multiple inputs: more \ac{lr} inputs better preserve identity and improve synthesis quality using early or late fusion. Furthermore, we add constraints to diversify the features (\ie orthogonal regularization) for more improvement.
\end{enumerate}

\begin{figure*}[t!] 
\centering   
\begin{tabular}{c}
	\centering    
    \begin{subfigure}[b]{0.9\linewidth}
     \centering
        \includegraphics[trim=0in 0.0in 0in 0in,clip,width=1\textwidth]{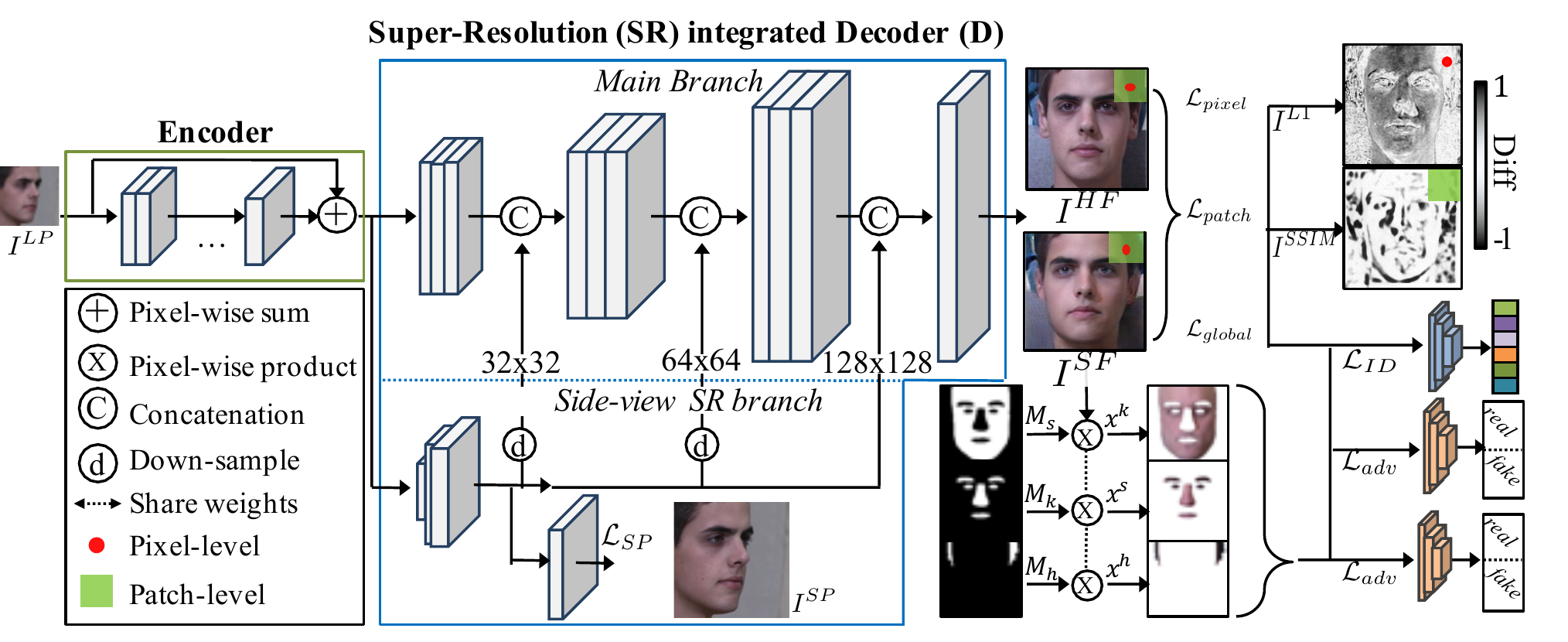}
         \caption{Proposed framework.}
         \label{fig:f1}
    \end{subfigure}\cr
    \begin{subfigure}[b]{0.8\linewidth}
     \centering
        \includegraphics[trim=0in 0in 0in 0in, clip, width=\textwidth]{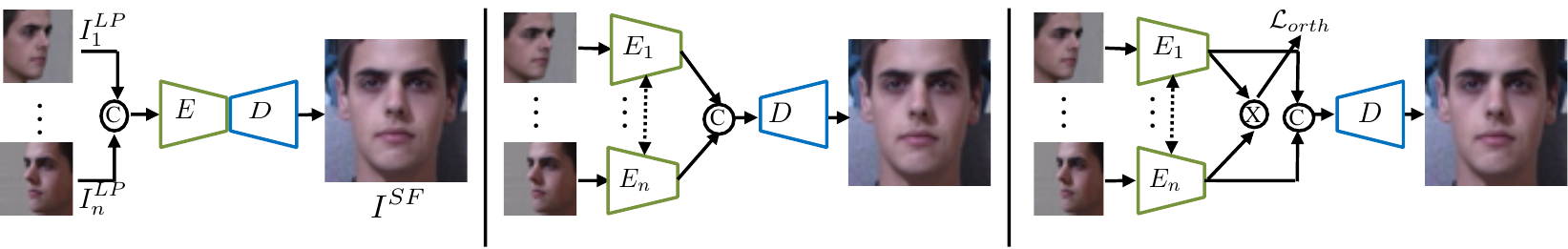}
         \caption{Fusion schemes.}
         \label{fig:f2}
    \end{subfigure}
\end{tabular}
    \caption{\textbf{Framework Overview.} (a) Given a non-frontal (\ie profile) LR face $I^{LP}$, SI \ac{sffgan} synthesizes a high-quality frontal face $I^{SF}$ by integrating a side-view SR. 
    (b) Furthermore, the proposed generalizes to multi-images of arbitrary poses as inputs-- each added sample improves results. Our MI \ac{sffgan}, even by naively fusing image inputs ((b) left), consistently outperforms SI in quality and identity preservation. Results are again boosted, drastically in fact (Table~\ref{tbl:ablationMI}), by fusing features after the encoder ((b) middle). Constraints force diverse features ((b) right); again, yielding a boost.}
         \label{fig:framework}
\end{figure*} 


\section{Related Work}

\subsection{Generative adversarial network}
Introduced in~\cite{goodfellow2014generative}, \ac{gan} train by facing \ac{g} off against \ac{d} in a {\em min-max} game, where \ac{g} aims to generate images indistinguishable from {\em real} $x$ from noise $z$ (\ie $G(z)\rightarrow \Tilde{x}$, where $\Tilde{x}$ is generated version of $x$). 
Recently, \ac{gan} have been successfully applied to various tasks like image-to-image translation~\cite{isola2017image}, image super-resolution~\cite{ledig2017photo}, and image in-painting~\cite{pathak2016context}. These successful applications of \ac{gan} motivate us to develop super-resolved frontal face synthesis methods based on \ac{gan}.





\subsection{Face frontalization}
Face frontalization is a challenging task due to incomplete information in face images when captured from a side-view. Previous attempts at the problem can be characterized in two fold: traditional (\ie {\em shallow}) methods and deep learning approaches. Traditional methods include \ac{3dmm} based methods~\cite{asthana2011fully,li2012morphable,koppen2018gaussian} and statistical-based models~\cite{tran2017disentangled,sagonas2015robust}. We focus the remainder of the literature review on the more relevant, state-of-the-art deep learning works~\cite{kan2014stacked,tran2017disentangled,yin2017towards,zhu2013deep,zhu2014multi,zhu2015high}.

Most similar to the proposed are GAN-based frontal-face synthesizers~\cite{donahue2016adversarial,huang2017beyond,tian2018cr,tran2017disentangled}.
BiGAN jointly learns \ac{g} and an inference model~\cite{donahue2016adversarial}. 
Nonetheless, in practice, BiGAN produces poor quality due to finite data and limited model capacity~\cite{tran2017disentangled}. 
Like us, DR-GAN~\cite{tran2017disentangled} learned identity-preserved representations to synthesize multi-view images. However, the encoder feeds the decoder, which depends on the training data-- an impractical restriction for the inability to generalize to new data.
TP-GAN has two pathways for frontal face generation to captured local and global features~\cite{huang2017beyond}. CR-GAN~\cite{tian2018cr} also had dual paths, with the addition of self-supervision to refine weights learned by the supervised module. We, too, look at various levels, including the addition of patch-level and enhanced global loss. 
FF-GAN~\cite{yin2017towards} adopted the \ac{3dmm} conditioned on a \ac{gan} as facial prior knowledge.
Finally, the work~\cite{li2019m2fpa} introduced the \ac{m2fpa}. Benchmarks for the new data included many of the \ac{gan}-based methods reviewed above. The authors also introduced an sufficient parsing guided \ac{d}, which we too incorporate in our model (Section~\ref{subsubsec:global}).


\subsection{Orthogonal regularization}
Orthogonal regularization forces the feature space to be more diverse. For this, some add a hard orthogonality constraint via singular value decomposition to remain on a Stiefel manifold~\cite{sun2017svdnet}. More recently, a {\em softer} variant was proposed, \ie orthogonality regularization via Gram matrix per weight matrix to force a near identity matrix by the spectral~\cite{2018arXiv181009102B} or Frobenius~\cite{xie2017all} norm: the former claims superiority with consistent improvements for \ac{cnn} with novel regularization scheme, Spectral Restricted Isometry Property (SRIP). SRIP proved to generalize well, and by an ease-of-use. Having showed such improvements in feed-forward \ac{cnn} trained for classification, we extend SRIP to a \ac{gan} (\ie \ac{sffgan}).

\section{Methodology}
We next define the face-frontalization problem from a single \ac{lr} image. Then, we describe the model and loss function of {\em \ac{si}} \ac{sffgan}. Finally, we introduce {\em \ac{mi}} \ac{sffgan} as an extension.

\subsection{Problem formulation}
Let $\{I^{HF}, I^{LP}\}$ be a pair of \ac{hr} frontal and \ac{lr} side-view faces. Given a side-view face $I^{LP}$, the goal is to train a generator \ac{g} to synthesize the corresponding HR frontal face image $\hat{I}^{HF}=G\left(I^{LP}\right)$ with identity-preserved in $I^{HF}$.

A depiction of the general architecture of the proposed \ac{sffgan} is in Fig.~\ref{fig:framework}. \ac{g} contains a deep encoder, a side-view SR module, and a decoder.
SR of side-view imagery is integrated into \ac{sffgan} to provide fine details of side-view faces, and hence help reconstruct higher frequency information (\ie periocular, nose, and mouth regions) of frontal faces.
Except for the novel architecture, we make this \ac{sr} and ill-posed problem well constrained by introducing a three-level loss (\ie pixel-, patch-, and global-based losses) that learns a precise non-linear transformation between LR side-view and HR frontal-view faces.

\subsection{Network architecture}
The generator \ac{g} contains a deep encoder, while the decoder contains a \ac{sr} module. Features extracted by the deep encoder are passed to the \ac{sr}-branch for reconstruction. The \ac{sr} side-view module feeds the decoder with higher frequency information to help reconstruct frontal faces. See supplemental material for all network specifications.

\noindent\textbf{Deep encoder.}
Previous works in face rotation often employ U-Net-like models~\cite{ronneberger2015u}. We argue that the usual encoder is too shallow to capture the high-frequency information needed to recover a high-quality, \ac{hr} face from \ac{lr} space. Instead, we adopt a deeper encoder to recover edges and shapes of the \ac{hr} frontal faces with higher precision. Another benefit is easing the task of \ac{sr} of a side-view \ac{lr} image, which provides details needed for reconstructing a \ac{hr} frontal face.
The encoder is shown in Fig.~\ref{fig:framework}: a $3\times3$ conv-layer followed by sixteen residual dense blocks \cite{zhang2018residual}. 

\noindent\textbf{SR-integrated decoder.}
From the output of the encoder, the two branches split, the side-view \ac{sr} module to super-resolve side-view images and, ultimately, pass feed back into the main path, along with the decoder that reconstructs \ac{hr} frontal faces (Fig.~\ref{fig:framework}). 
The {\em side-view \ac{sr}} uses pixel shuffle to up-sample \cite{shi2016real} by a factor of 4 (\ie $128\times128$). From this, higher frequency content is fed to the main branch at various sizes to help reconstruct frontal faces.

\begin{figure*}[!t]
	\centering   
    \includegraphics[trim=0in 1.32in 0in 0in,clip,width=0.8\linewidth]{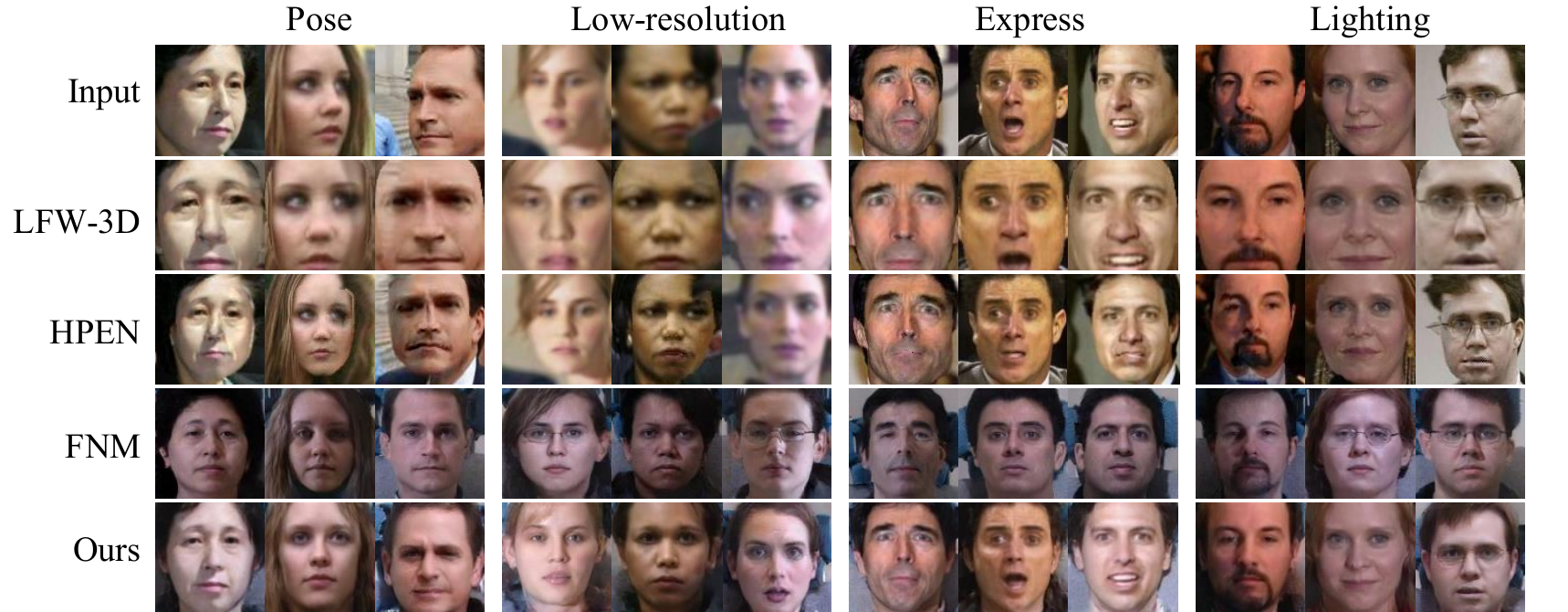}
    \caption{\textbf{Qualitative results on LFW~\cite{LFWTech}.} Comparison with SOTA under extreme pose, LR, expression, and lighting.}\label{fig:compare_lfw}
\end{figure*}

\subsection{Loss function}

\noindent\textbf{Pixel-level loss.}
L1 loss is used as a pixel-level loss, since it provides better convergence than L2 in supervised image generation tasks. We adopt pixel-wise L1 loss to measure both super-resolved side-view faces $I^{SP}$ and synthesized frontal faces $I^{SF}$:
 \begin{equation}
     \mathcal{L}_{pix}  = \frac{1}{W \times H} \sum_{w, h=1}^{W, H}\left| I_{w,h}^{HP}-I_{w,h}^{SP}\right| + \left| I_{w,h}^{HF}-I_{w,h}^{SF}\right| ,
 \end{equation}
where $W$ and $H$ are the width and height of synthesized images (\ie 128$\times$128), respectively. $I^{HP}$ and $I^{HF}$ denote HR side-view and HR frontal faces, respectively.

\noindent\textbf{Patch-level loss.}
Comparing to pixel-level loss, patch-level loss pays more attention to image structures (\ie the edge and shape of facial components). Here we adopted \ac{ssim} \cite{wang2004image} as patch-level loss to capture structural information and compliment pixel-level loss.
\ac{ssim} measures the perceptual difference between a generated and a reference image. Let $\mathbf{x}=\{x_1,...,x_{K^2}\}$ and $\mathbf{y}=\{y_1,...,y_{K^2}\}$ be the pixel values of two corresponding $K\times K$ patches cropped from the synthesized $I^{SF}$ and the HR frontal face $I^{HF}$, respectively. The SSIM of $\mathbf{x}$ and $\mathbf{y}$ is computed as
\begin{equation}
    SSIM(\mathbf{x},\mathbf{y}) = 1 - \frac{(2\mu_x\mu_y+C_1)(2\sigma_{xy}+C_2)}{(\mu_x^2\mu_y^2+C_1)(\sigma_x^2+\sigma_y^2+C_2)},
\end{equation}
where $\mu_x$, $\mu_y$ and $\sigma_x$, $\sigma_y$ corresponds to the mean and standard deviation of $\mathbf{x}$ and $\mathbf{y}$, respectively. And $\sigma_{xy}$ is the covariance of $\mathbf{x}$ and $\mathbf{y}$. Constraints $C_1=0.01^2$ and $C_2=0.03^2$ are added for numeric stability.



Then, the patch-level loss is defined over $P$ patches as

\begin{equation}
    \mathcal{L}_{patch}  = \frac{1}{P}\sum_{p=1}^P SSIM(\mathbf{x}^p, \mathbf{y}^p).
\end{equation}

\noindent\textbf{Global-level loss.}\label{subsubsec:global}
In the global-level are adversarial and identity-preserving losses to synthesize photo-realistic frontal faces with high-frequency details and consistent identity as the input.

\textit{Adversarial loss.}
The frontal-face generative models should pay attention to all details used to distinguished a face as a whole to synthesize photo-realistic, frontal faces. Inspired by \cite{li2019m2fpa}, we employ two discriminators at training (\ie one for frontal faces $D_f$ and another parsing-guided $D_p$). $D_f$ aims to distinguish real \ac{hr} frontal faces $I^{f}$ from synthesized $\hat{I}^{f}$. $D_p$, although aims to work with \ac{d}$_f$, focuses on different facial regions. Specifically, a pre-trained face parsing model~\cite{liu2015multi} to generates images regions $I^p$ to capture low-frequency information (\ie skin regions), key-points (\ie eyes, brows, nose, and lips), and hairline as 
\begin{align}\label{eq:masked}
&{\textit{real}} ~~I^p= \{I^f{\odot} M_s, I^f{\odot} M_k,  I^f{\odot} M_h\},\nonumber\\
&{\textit{fake}} ~~\hat{I}^p=\{\hat{I}^f{\odot} M_s, \hat{I}^f{\odot} M_k, \hat{I}^f{\odot} M_h \}\text{.}
\end{align}
where $M_s, M_k, M_h$ are skin, key-points, and hairline masks (Fig.~\ref{fig:framework}). ${\odot}$ is the element-wise product.

Then, the overall adversarial loss can be expressed as
\begin{eqnarray}  
\begin{aligned}
    \mathcal{L}_{adv}=&\sum_{j\in\{f, p\}}\Big (\mathbb{E}_{I^j}\left[\log D_j(I^j)\right]\\
    &~~~~~~~~~~~~~~+\mathbb{E}_{\hat{I}^j}[\log(1 - D_j(\hat{I}^j))] \Big).
\end{aligned}
\end{eqnarray}

\textit{Identity preserving loss.}
A critical aspect of evaluating face frontalization is the preservation of identities during the synthesis of frontal faces. We exploit the ability of pre-trained face recognition networks to extract meaningful feature representations to improve the identity preserving ability of \ac{g}. Specifically, we employ a pre-trained 29-layer Light CNN\footnote{Downloaded from \href{https://github.com/AlfredXiangWu/LightCNN}{https://github.com/AlfredXiangWu/LightCNN}.}~\cite{wu2018light} with its weights fixed during training to calculate an identity preserving loss for \ac{g}. The identity preserving loss is defined as the feature-level difference in the last two fully connected layers of Light CNN between the synthesized image $I^{SF}$ and the ground-truth $I^{HF}$:
 \begin{equation}
    \mathcal{L}_{ID} = \sum_{i=1}^2 ||p_i(I^{SF})-p_i(I^{HF})||^2_2
 \end{equation}
where $p_i(i\in{1,2})$ denotes the outputs of the two fully connected layers of LightCNN, and $||\cdot||_2$ denotes the L2-norm.

\noindent\textbf{Overall loss.}
The objective function for the proposed is a weighted sum of aforementioned three-level losses:
\begin{equation}
    \mathcal{L}_{G} = \lambda_1\mathcal{L}_{pix} +\lambda_2\mathcal{L}_{patch} + \lambda_3\mathcal{L}_{adv} +  + \lambda_4\mathcal{L}_{ID} + \lambda_5\mathcal{L}_{tv},
\end{equation}
where $\lambda_1$, $\lambda_2$, $\lambda_3$, $\lambda_4$, and $\lambda_5$ are hyper-parameters that control the trade-off of the loss terms. A total variation regularization $\mathcal{L}_{tv}$ \cite{johnson2016perceptual} is also included to remove unfavorable artifacts in synthesized frontal faces $I^{SF}$.

\subsection{Multi-image \ac{sffgan}}
\ac{si} \ac{sffgan} synthesized a \ac{sr} frontal face from one side-view image. Yet, we often have multiple images per subject in real-life scenario (\eg surveillance system). To leverage the complimentary information of different poses, we propose \ac{mi} \ac{sffgan} that can penalize redundant latent representations and explore the maximum information of the \ac{lr} images under arbitrary poses. To be specific, \ac{mi} \ac{sffgan} use the same decoder as SI \ac{sffgan}, but multiple encoders with shared weights for different input images. The framework of \ac{mi} \ac{sffgan} is shown in Fig.~\ref{fig:f2}. Different from image-level and feature-level fusion, \ac{mi} \ac{sffgan} introduce a constrain (\ie orthogonal regularization) on the features extracted from the encoder. The orthogonal constrain makes the features more diverse and hence compliment each other as much as possible. We augment the objective function of \ac{si} \ac{sffgan} with the loss:
\begin{equation}
    \mathcal{L}_{orth}  = \frac{1}{N}\sum_{n=1}^N || G(I^{LP}_n)^\top G(I^{LP}_n)||^2_F,
\end{equation}
where $N$ is the number of LR input images. $||\cdot||^2_F$ denotes the squared Frobenius norm.
Then the loss function for \ac{mi} \ac{sffgan} can be expressed as:
\begin{equation}
    \mathcal{L}_{MI}  = \mathcal{L}_{G} + \mathcal{L}_{orth}.
\end{equation}

\section{Experiment}
We now demonstrate the effectiveness of \ac{sffgan} for \ac{hr} frontal face synthesis and pose-invariant representation learning. We show quantitative synthesis results of the \ac{si} and \ac{mi} \ac{sffgan} and compare with the state-of-the-art methods trained on both \ac{lr} and \ac{sr} inputs. Besides, we highlight the identity preserved by the proposed by quantitatively evaluate the face recognition performance. Finally, we do an ablation study as a deep-dive revealing the benefits of the SR-integrated architecture, different types of loss function,
and the multi-view fusion method.

\subsection{Settings}
\noindent\textbf{Datasets.}\label{sec:Dataset} 
We conduct experiments on the Multi-PIE~\cite{gross2010multi} and CAS-PEAL-R1~\cite{gao2007cas} datasets. The Multi-PIE consists of 337 subjects involved in up to 4 sessions. Each session included 20 illumination levels, 15 poses (\ie within $\pm90^\circ$, with a step size of $\pm15^\circ$), and six expressions (\ie neutral, smile, surprise, squint, disgust, and scream). Per convention, two settings of Multi-PIE followed \cite{yim2015rotating,huang2017beyond,hu2018pose,li2019m2fpa}. 

\emph{Setting 1} emphasizes pose, illumination, and minor expression variations. Thus, we only include a single image sample per session (\ie \textit{Session 1}). There are 250 identities, with the first 150 set as the training set, which includes the 9 poses spanning $\pm$60$^\circ$ and 20 illumination levels per subject. A frontal face, neutral in expression and illumination, of the other 100 identities makes-up the search gallery, with the remaining face samples of these subjects set as probes. 

\emph{Setting 2} emphasizes pose, illumination, and session (\ie time) variations. Samples contain faces with neutral expressions from all four sessions and of all 337 identities. Samples 
of the first 200 identities are set as the training set; the remaining 137 subjects are used for testing, with, again, samples neutral in expression and illumination as the gallery. 
In total, there are 161,460, 72,000, and 137 faces for training, as test probes, and in the gallery, respectively. Notice, no overlap in subjects between train and test.

The CAS-PEAL-R1 dataset is a public large-scale face database made-up of pose, expression, accessory, and lighting variations. The dataset contains 30,863 grayscale pose images of 1,040 subjects, 595 males and 445 females, with 7 yaw angles within $\pm$45$^\circ$ and 3 pitch angles within $\pm$30$^\circ$, totaling to 21 yaw-pitch rotations. The training set is of all images for the first 600 subjects, and 440 for testing.

\begin{figure}[!t]
	\centering   
    \includegraphics[trim=0in 0.36in 3.5in 0in,clip,width=0.88\linewidth]{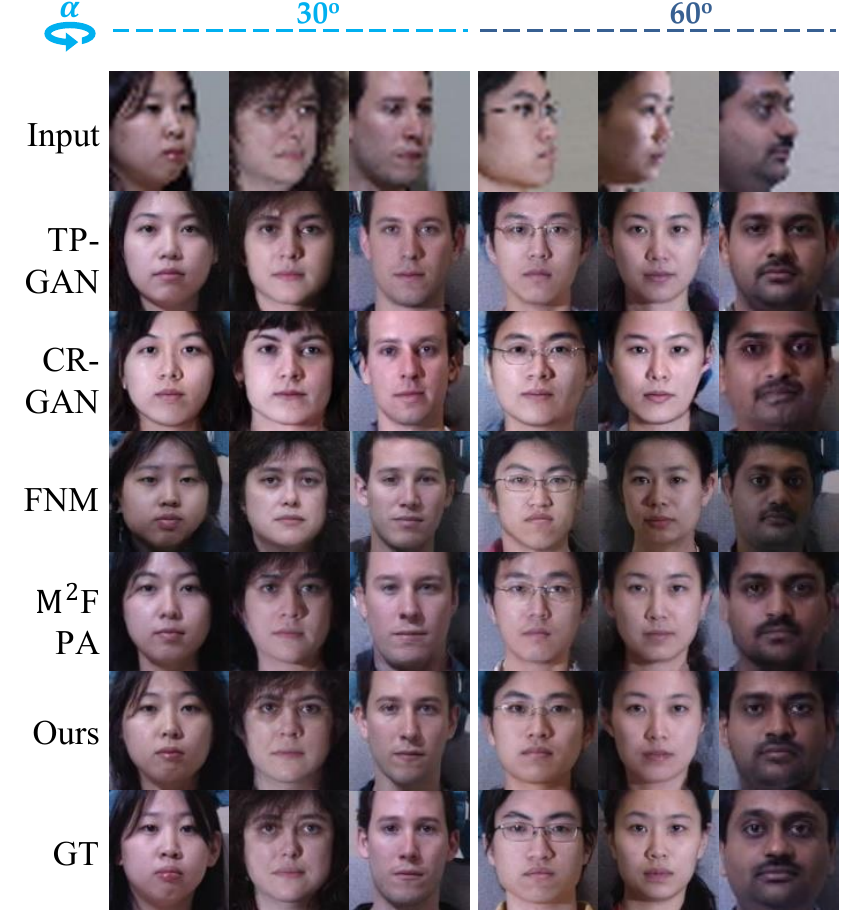}
    \caption{\textbf{Qualitative results on Multi-PIE.} }\label{fig:compare_state_of_the_art}
\end{figure}

LFW~\cite{huang2008labeled} contains 13,233 face images collected in unconstrained environment. It will be used to evaluate the frontalization performance in uncontrolled settings.

\begin{table*}[t]
    \centering
    \caption{\textbf{Multi-PIE Setting 2.} PSNR (dB), SSIM and Rank-1 (\%) performance across views ($\alpha$). \ac{si} \ac{sffgan} on \ac{lr} input can recover even better identity information and finer detail than the state-of-the-art methods trained on super-resolved input. }
    \begin{adjustbox}{max width=\linewidth}
    \begin{tabular}{lrccccccccccccccccc}
    \toprule
    & &\multicolumn{5}{c}{$\bf{PSNR}$} && \multicolumn{5}{c}{\textbf{$\bf{SSIM}$}} && \multicolumn{5}{c}{$\bf{Rank-1}$} \\
    \cline{2-7}    \cline{9-13}    \cline{15-19} 
    &$\mathbf{\alpha}$& $\bf{\pm 60^\circ}$ & $\bf{\pm 45^\circ}$ & $\bf{\pm 30^\circ}$ & $\bf{\pm 15^\circ}$ & \textbf{Avg} && $\bf{\pm 60^\circ}$ & $\bf{\pm 45^\circ}$ & $\bf{\pm 30^\circ}$ & $\bf{\pm 15^\circ}$ & \textbf{Avg} &&  $\bf{\pm 60^\circ}$ & $\bf{\pm 45^\circ}$ & $\bf{\pm 30^\circ}$ &$\bf{\pm 15^\circ}$ & \textbf{Avg} \\
    \midrule
    \multirow{2}{*}{TP-GAN~\cite{huang2017beyond}}&\textbf{LR}& 19.00 & 19.22 & 19.52 & 19.67 & 19.35 && 0.625 & 0.634 & 0.645 & 0.654 & 0.640 &&  57.87& 65.78 & 69.13 & 73.99 & 66.69 \\ 
    \multirow{2}{*}{} &\textbf{SR}& 19.06 & 19.32 & 19.55 & 19.69 & 19.41 && 0.638 & 0.647 & 0.657 & 0.664 & 0.652 &&  68.95& 77.58 & 81.69 & 84.83 & 78.26 \\
    \cdashline{2-19}[1pt/1pt]

    \multirow{2}{*}{CR-GAN~\cite{tian2018cr}}&\textbf{LR}& 16.83 & 18.04 & 18.03 & 18.76 & 17.92 && 0.499& 0.536& 0.548& 0.567 & 0.538 &&  46.89& 57.53 & 60.93 & 65.54 & 57.72 \\ 
    \multirow{2}{*}{} &\textbf{SR}& 19.53& 19.71& 19.92& 20.20 & 19.84 && 0.632& 0.640& 0.648& 0.653 & 0.643 &&  67.45& 70.63 & 71.44 & 71.99 & 70.38 \\
    \cdashline{2-19}[1pt/1pt]

    \multirow{2}{*}{FNM~\cite{qian2019unsupervised}}&\textbf{LR}& 15.50& 15.79& 16.32& 17.25 & 16.22 && 0.433& 0.439& 0.451& 0.470 &0.448 &&  62.68& 66.54 &69.22 & 72.31 & 67.69\\ 
    \multirow{2}{*}{} &\textbf{SR}& 15.58& 16.00& 16.72& 17.61 & 16.48 && 0.427& 0.436& 0.449& 0.468 & 0.445 &&  76.21& 79.46 &82.08 & 85.05 & 80.7  \\   
    \cdashline{2-19}[1pt/1pt]
    
    \multirow{2}{*}{M2FPA~\cite{li2019m2fpa}}&\textbf{LR}& 22.38& 22.73& 23.17& 23.91 & 23.05 && 0.692& 0.704& 0.719& 0.743 & 0.715 &&  66.50& 76.95 &84.24 & 90.52 & 79.55 \\ 
    \multirow{2}{*}{} &\textbf{SR}& 22.48& 22.82& 23.29& 24.11 & 23.18 && 0.697& 0.710& 0.728& 0.755 & 0.723 &&  79.32& 88.03 &93.35 & 97.63 & 89.58  \\   
    \midrule
  
    \ac{si} \ac{sffgan}  &\textbf{LR} &\textbf{22.58}  & \textbf{23.02}& \textbf{23.56}&  \textbf{24.53}  & \textbf{23.42} && \textbf{0.700} &\textbf{0.717} &\textbf{0.736} & \textbf{0.764} &\textbf{0.729} && \textbf{85.25}&\textbf{92.31}&\textbf{95.85} & \textbf{97.82}& \textbf{92.81}\\
    \bottomrule
    \end{tabular}\label{tbl:setting2_all} 
    \end{adjustbox}
\end{table*}

\noindent\textbf{Implementation.}
Training requires images pairs $\{I^{LP}, I^{HF}\}$, one LR side-view image and the corresponding HR frontal face. We first cropped images to a canonical view (128$\times$128), making-up the \ac{hr} images~\cite{huang2017beyond}. Then, \ac{lr} images are created by bicubic downsampling (4$\times$). Unlike CAS-PEAL-R1, Multi-PIE is RGB. Thus, the identity-preserving model for Multi-Pie and CAS-PEAL-R1 were pre-trained on MS-Celeb-1M~\cite{guo2016ms} in RGB and gray-scale, respectfully, and then fine-tuned on the respective training set.
We implemented the model in PyTorch. Parameters were set as follows: $\lambda_{1}=20$, $\lambda_{2}=5$, $\lambda_3=0.8$, $\lambda_4=0.1$, $\lambda_4=1^{-4}$, $\lambda_3=0.1$.
We used an ADAM optimizer with a learning rate of $10^{-4}$, reducing 0.5 at 10$^{th}$ and 15$^{th}$ epochs for SI, and 5$^{th}$ and 10$^{th}$ epochs for MI (\ie faster convergence). Batch of 8 for 20 epochs.

\subsection{Face synthesis}
In this section, we show results of the \ac{si} and \ac{mi} \ac{sffgan}, and compare with state-of-the-art methods. We used the public code of TP-GAN\footnote{\href{https://github.com/HRLTY/TP-GAN}{https://github.com/HRLTY/TP-GAN}.}~\cite{huang2017beyond}, CR-GAN\footnote{\href{https://github.com/bluer555/CR-GAN}{https://github.com/bluer555/CR-GAN}.}~\cite{tian2018cr}, and re-implemented M$^2$FPA~\cite{li2019m2fpa} since code is not public. 
For Multi-PIE, all the models were trained and evaluated for both settings (Section~\ref{sec:Dataset}). Note that all synthesized results in paper were generated under {\em setting 2}.
Fig.~\ref{fig:compare_state_of_the_art} (a) shows a qualitative comparison on Multi-PIE and demonstrates the superior performance of the proposed \ac{sffgan} on \ac{lr} images. Qualitative results show that the proposed \ac{sffgan} can recover HR frontal faces from LR side-views with identity preserved and finer details (\ie more precise facial shapes and textures). Notice other methods produce frontal faces with more inaccuracies due to blurry input.

We then demonstrate the robustness of \ac{sffgan} to extreme pose, low-resolution, expression, and lighting. Fig.~\ref{fig:compare_lfw} shows the comparison results of \ac{sffgan} and state-of-the-art methods (\ie LFW-3D~\cite{hassner2015effective}, HPEN~\cite{zhu2015high}, and FNM~\cite{qian2019unsupervised}) on the unconstrained dataset LFW~\cite{huang2008labeled}. More results on LFW will be provided in supplementary material.

\begin{figure*}[t!]
    \centering
    \includegraphics[trim=0in 2.05in 0in 0in,clip,width=\linewidth]{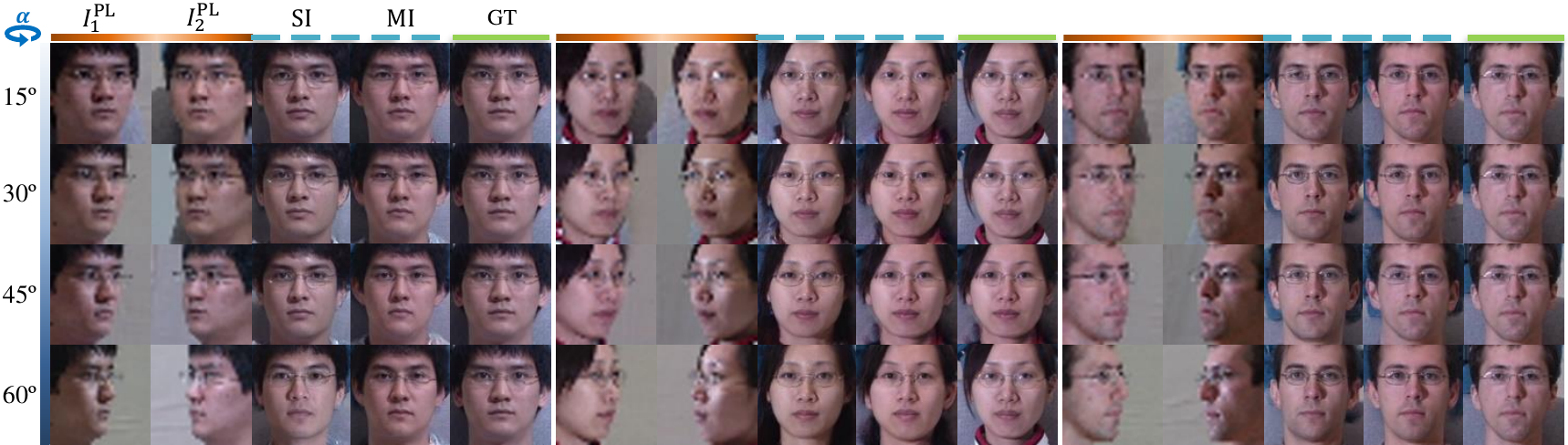}
    \caption{\textbf{\ac{si} and \ac{mi} \ac{sffgan} synthesis results.} \ac{si} SSF-GAN recovers better frontal faces than existing methods for different yaws ($\alpha$) (Fig.~\ref{fig:compare_state_of_the_art}). However, \ac{mi} \ac{sffgan} further improves the image quality and identity preserving ability.}
    \label{fig:facemontage}
\end{figure*}

\begin{table}[t]
    \centering
    \caption{\textbf{CAS-PEAL-R1.} Rank-1 recognition performance (\%) across pitches ($\beta$).}
    \begin{adjustbox}{max width=0.7\linewidth}
    \begin{tabular}{lrccccc}
    \toprule
    ~~~~~~~~~~$\beta$&&{$\bf{-15^\circ}$} && {\textbf{$\bf{0^\circ}$}} && {$\bf{+15^\circ}$} \\

    \midrule
    \multirow{2}{*}{TP-GAN~\cite{huang2017beyond}}&\textbf{LR} &  90.97 &&  94.96 &&  91.01\\ 
    \multirow{2}{*}{} &\textbf{SR}  &94.08 &&  97.70 && 94.50 \\
    \cdashline{2-7}[1pt/1pt]
    
     \multirow{2}{*}{CR-GAN~\cite{tian2018cr}}&\textbf{LR} &72.86 &&  87.92 &&  78.94\\ 
    \multirow{2}{*}{} &\textbf{SR} & 79.51 &&  89.80 && 84.45\\
    \cdashline{2-7}[1pt/1pt]
    
    \multirow{2}{*}{M2FPA~\cite{li2019m2fpa}}&\textbf{LR} & 94.36 &&  98.21&&  96.32\\ 
    \multirow{2}{*}{} &\textbf{SR} & 97.91  && 99.35&&  98.74\\
    \midrule
    
    \ac{si} \ac{sffgan}&\textbf{LR}  & \textbf{98.06} &&  \textbf{99.88} &&\textbf{98.87}\\
    \bottomrule
    \end{tabular}\label{tbl:rank1_cas} 
    \end{adjustbox}
\end{table}


Considering the other methods are designed for \ac{hr} input and tend to fail learning the highly non-linear representation from \ac{lr} side-view to \ac{hr} frontal faces. We then quantitatively compare synthesis results of the state-of-the-art methods generated from super-resolved side-view faces for more fair comparison (see Table \ref{tbl:setting2_all}). We employ a pre-trained RCAN\footnote{\href{https://github.com/yulunzhang/RCAN}{https://github.com/yulunzhang/RCAN}.}~\cite{zhang2018image} as image SR model to generate super-resolved side-view images.
Results are reported on PSNR and SSIM (Table \ref{tbl:setting2_all}) of synthesized frontal face. Quantitative results show that the proposed SI \ac{sffgan} can not only achieve better results on LR input, but still recover frontal faces with better quality and finer structure than the state-of-the-art methods trained on SR input.

Fig.~\ref{fig:facemontage} shows the synthesized \ac{hr} frontal results of both \ac{si} and \ac{mi} \ac{sffgan} with poses of $15^\circ$, $30^\circ$, $45^\circ$, $60^\circ$. Notice, photo-realistic faces are synthesized from one-to-many LR inputs of arbitrary views. The results for \ac{mi} \ac{sffgan} were from two \ac{lr} inputs: the one used for the \ac{si}, and the other the inverted counterpart (\ie $\pm15^\circ$, $\pm30^\circ$, $\pm45^\circ$, $\pm60^\circ$). 
Note that all synthesized results of \ac{si} \ac{sffgan} are consistent with the ground-truth (GT) faces, showing clear superiority across the different pose and lighting variations. Moreover, \ac{mi} \ac{sffgan} further improves the image quality of the synthesized images, while preserving the identity even better than \ac{si} \ac{sffgan}.


\subsection{Identity preserving property}\label{sec:experiment_identity} 
To quantitatively demonstrate the identity preserving ability of proposed \ac{sffgan}, we evaluate face recognition accuracy on synthesized frontal images. Table~\ref{tbl:setting2_all} compares face recognition performance with existing state-of-the-art on setting 2 of Multi-PIE across different poses. Results on setting 1 of Multi-PIE are shown in the supplementary material. Results are reported with Rank-1 recognition accuracy. We conduct the experiment by extracting features using a pre-trained face recognition model (\ie 29-layer Light-CNN\cite{wu2018light}), and then compute feature similarities via cosine-distance metric. Results on setting 2 of Multi-PIE shows that \ac{sffgan} consistently achieves the best performance across all angles. 
Note that the existing methods tend to fail capturing identity information from LR input, while the performance is largely improved with a two-step processing, which is to first super-resolve the side-view faces and then frontalize them. However, the proposed \ac{sffgan} can recover identity preserving and HR frontal faces directly from LR images, and have even better performance than the SOTA methods trained with SR images. 

We analyze, quantitatively, the benefits of using the proposed in the LFW benchmark (Table~\ref{tab:lfw_face_recognition}).
Specifically, face recognition performance is evaluated on synthesized frontal images. The results of SOTA are in Table~\ref{tab:lfw_face_recognition} are from \cite{li2019m2fpa}.  Similarly, we show the rank-1 recognition accuracy for CAS-PEAL-R1 across pitch ($\beta$) pose variations. The quantitative results are summarized in Table~\ref{tbl:rank1_cas}, which demonstrates that \ac{sffgan} significantly outperforms its competitors in terms of identity preservation.

To the best of our knowledge, only DR-GAN~\cite{tran2017disentangled} attempted to solve the MI frontal face problem. We follow the settings in~\cite{tran2017disentangled} for MI fusion. First, a subset $\mathbb{P}_0$ of images with poses in (30$^\circ$, 60$^\circ$) is selected from the Multi-PIE probe set. Then, we form four probe sets $\{\mathbb{P}_i\}^4_{i=1}$ with the image count ranging from 1-to-4. Specifically, $\mathbb{P}_1$ is formed by randomly selecting one image per subject from $\mathbb{P}_0$. $\mathbb{P}_2$ is formed by adding a random image per subject to $\mathbb{P}_1$.
Similarly, $\mathbb{P}_3$ and $\mathbb{P}_4$ are constructed.
For face recognition, we directly obtained results reported in the DR-GAN paper, since we fail training the model on LR and SR images. Note that results for DR-GAN are trained and tested on HR images, with ours outperforming most (Table~\ref{tbl:ablationMI}).

\begin{table}
    \centering  
    \caption{\textbf{LFW benchmark.} Face verification accuracy (ACC) and area-under-curve (AUC) results.}
    \begin{adjustbox}{max width=0.65\linewidth}
    \begin{tabular}{c|c|c}
    \hline
      & ACC (\%) & AUC (\%)\\
    \hline
    LFW-3D~\cite{hassner2015effective} & 93.62   & 88.36 \\
    LFW-HPEN~\cite{zhu2015high} & 96.25  & 99.39 \\
    FF-GAN~\cite{yin2017towards} & 96.42   & 99.45 \\
    CAPG-GAN~\cite{hu2018pose} & 99.37   & 99.90 \\
    M$^2$FPA~\cite{li2019m2fpa} & 99.41   & 99.92 \\
    FNM~\cite{qian2019unsupervised} & 99.42   & 99.93 \\
    \hline
    Ours &  99.48  & 99.96\\
    \hline
    \end{tabular}
    \label{tab:lfw_face_recognition}
    \end{adjustbox}
\end{table}

\setlength{\tabcolsep}{4pt}
\begin{table}[t]
    \centering
    \caption{\textbf{Ablation Study of SI and MI \ac{sffgan}.} Average Rank-1 (\%), PSNR (dB) and SSIM on MutiPIE setting 2.}
    \small
    \begin{adjustbox}{max width=0.75\linewidth}
    \begin{tabular}{c|cccc}
    \toprule
    && \ Rank-1\ \  & \ PSNR\ \  & \ SSIM\ \ \\
    \midrule
    \parbox[t]{5mm}{\multirow{7}{*}{{\vspace{-0.14in}SI}}} &baseline\_1 & 84.99 & 23.21 & 0.718\\ 
    \multirow{7}{*}{} & baseline\_2 & 91.34 & 23.38 & 0.724 \\ 
  \cdashline{2-5}[1pt/1pt]
    \multirow{7}{*}{} &w/o L1 (pixel) & 90.81 & 23.34 & 0.730 \\ 
    \multirow{7}{*}{} &w/o \ac{ssim} (patch) & 91.16 & 23.29 & 0.727 \\ 
    \multirow{7}{*}{} &w/o ID (global) & 81.49 & 23.32 & 0.726 \\ 
    \multirow{7}{*}{} &w/o Adv (global) & 91.81 & \textbf{23.87} & \textbf{0.758} \\ 
  \cdashline{2-5}[1pt/1pt]
    \multirow{7}{*}{} &\ac{sffgan} & \textbf{92.81} & 23.42 & 0.729 \\
  \midrule
    \parbox[t]{5mm}{\multirow{3}{*}{{\vspace{-0.04in}MI}}}&image-level & 96.09 & 23.98  & 0.755 \\ 
    \multirow{3}{*}{}&feat-level & 97.68 & 24.21 & 0.760 \\ 
    \multirow{3}{*}{}&\ac{sffgan} & \textbf{98.43} & \textbf{24.43} & \textbf{0.762} \\ 
  \bottomrule
    \end{tabular}\label{tbl:ablation_SI} 
    \end{adjustbox}
\end{table}

\subsection{Ablation study}
We conduct an ablation study as a deep-dive revealing the benefits of the SR-integrated architecture, the different synthesis loss function, and the multi-view fusion method. 

\textbf{Effect of \ac{sr} side-view module.}
To highlight the importance of {\em SR side-view}, we compare \ac{sffgan} with and without the SR module (Table \ref{tbl:ablation_SI}). Specifically, we remove the \ac{sr} module (\ie { \em baseline\_1}). Then, the same structure as \ac{sffgan} except with no supervision for {\em SR side-view} (\ie {\em baseline\_2}). We use Rank-1, PSNR and \ac{ssim} to evaluate.
Results show {\em baseline\_1} performs the poorest, implying the second branch learns complimentary features even without supervision. Hence, the added high-frequency information for frontal face synthesis persists. Moreover, the performance is further improved with a supervised {\em SR side-view}, validating the contributions of this module.

\setlength{\tabcolsep}{5pt}
\begin{table}[t]
    \centering
    \caption{\textbf{\ac{mi} performance.} Rank-1 recognition performance (\%) of different fusion scheme (\ie image-level, feature-level, and with the proposed constraint in \ac{g}).}
    \begin{adjustbox}{max width=0.75\linewidth}
    \begin{tabular}{ccccc}
    \toprule
    \# of images & $\bf{1}$ & $\bf{2}$ & $\bf{3}$ & $\bf{4}$\\
    \midrule
    DR-GAN~\cite{tran2017disentangled}& 85.90& \textbf{92.80}& 95.10 & 96.00\\
    \cdashline{1-5}[.4pt/1pt]
        image-level & \textbf{91.14} & 87.14& 90.71 & 93.69 \\ 
    feat-level & \textbf{91.14} & 92.01 & 96.80 & 99.03\\
  +constraints& \textbf{91.14} & 92.74 &\textbf{96.97}& \textbf{99.19} \\ 
    \bottomrule
    \end{tabular}\label{tbl:ablationMI} 
    \end{adjustbox}
    \vspace{-3mm}
\end{table}

\textbf{Effect of different losses.}
We show the contribution of each type of loss
by removing one of the three losses, pixel- (\ie $L_1$), patch- (\ie $L_{SSIM}$), or global- (\ie $L_{ID}$, $L_{Adv}$) (Table \ref{tbl:ablation_SI}). Rank-1 recognition accuracy, PSNR, and \ac{ssim} of frontalized images are used for evaluation. The qualitative results are shown in the supplementary material. We observe the recognition performance sharply decreases with the global and without the pixel. Though the PSNR and \ac{ssim} is largely improved without $L_{Adv}$, the synthesized images tend to be blurry without $L_{Adv}$. 
Comparing results between patch ($L_{SSIM}$) and pixel-level ($L_1$) loss, we observed that recognition accuracy is higher without $L_{SSIM}$, while PSNR and \ac{ssim} are higher without $L_1$. With both loss, the proposed \ac{sffgan} achieves the best in terms of Rank-1 and PSNR, and comparable in SSIM (only 0.001 lower), implying that $L_{SSIM}$ and $L_1$ are complimentary.

\textbf{Effect of fusion scheme via orthogonal constrain.}
We gain insight by exploring various fusion methods. We conduct two experiments to validate the proposed fusion scheme: (1) fuse two LR images with opposite poses $\alpha$ and $-\alpha$ (Table \ref{tbl:ablation_SI}), and (2) fuse \ac{mi} with arbitrary poses spanning [30$^\circ$, 60$^\circ$] (Table \ref{tbl:ablationMI}).
For \ac{mi} fusion, we follow the experiment setting of DR-GAN (Section~\ref{sec:experiment_identity}). We compare Rank-1 recognition accuracy and PSNR of synthesized images for both.
The results demonstrate that more inputs improve identity preservation when fused in image or feature space. Moreover, we gain further improvement by adding orthogonal constraints to penalize redundant latent representations and diversify the features.

\section{Conclusion}\acresetall
    We proposed SuperFront generative adversarial network (SF-GAN) to synthesize photo-realistic, identity-preserving frontal faces from LR-to-HR. To the best of our knowledge, we are the first to address the problem of rotating tiny profile faces. Our {\em SR side-view} module enhances faces to provide the high-frequency details needed to produce high-quality, identity-preserving faces. 
    In parallel to this, we introduce a patch loss that captures facial structure information and complements pixel loss. 
    With the two modules fused, a precise non-linear mapping between LR side-view and HR frontal-view is learned. Furthermore, the proposed model handles single and multi-image inputs-- more samples with arbitrary poses per subject as input, the better the quality of the synthesized output. A constraint is imposed on multi-image inputs to remove redundant information (\ie orthogonal regularization). We explore different fusion techniques, providing an ablation study to characterize our model in a complete, transparent manner. Quantitative and qualitative results demonstrate SF-GAN as state-of-the-art.

\clearpage

{\small
\bibliographystyle{ieee_fullname}
\bibliography{egbib}
}

\end{document}